\documentclass[a4paper,10pt,twoside]{article}

\usepackage{clin}        
\usepackage{harvard}     
\usepackage{tikz}
\usetikzlibrary{shapes.geometric, arrows}
\usepackage{graphicx}
\usepackage{paralist}
\usepackage{url}
\usepackage{here}
\usepackage[utf8]{inputenc}
\usepackage{float}
\usepackage{multirow}
\usepackage{subcaption}
\usepackage{makecell}
\usepackage{color}

\begin{document}

\title{FullStop: Punctuation and Segmentation Prediction for Dutch with Transformers}

\author{Vincent Vandeghinste$^*$ \email{vincent.vandeghinste@ivdnt.org}\\
{\normalsize \bf Oliver Guhr}$^\dagger$ \email{oliver.guhr@htw-dresden.de}\\
\AND \addr{$^{*}$Instituut voor de Nederlandse Taal, Leiden, the Netherlands and Centre for Computational Linguistics, Leuven.AI, KU Leuven, Belgium}
\AND \addr{$^\dagger$Hochschule für Technik und Wirtschaft, Dresden, Germany}}

\maketitle\thispagestyle{empty} 


\begin{abstract}
When applying automated speech recognition (ASR) for Belgian Dutch \cite{ASR}, the output consists of an unsegmented stream of words, without any punctuation. A next step is to perform segmentation and insert punctuation, making the ASR output more readable and 
easy to manually correct. As far as we know there is no publicly available punctuation insertion system for Dutch that functions at a usable level.

The model we present here is an extension of the models of \citeasnoun{sepp} for Dutch and is made publicly available.\footnote{\url{ https://huggingface.co/oliverguhr/}}  We trained a sequence classification model, based on the Dutch language model RobBERT \cite{delobelle-etal-2020-robbert}. For every word in the input sequence, the models predicts a punctuation marker that follows the word. We have also extended a multilingual model, for cases where the language is unknown or where code switching applies. 

When performing the task of segmentation, the application of the best models onto out of domain test data, a sliding window of 200 words of the ASR output stream is sent to the classifier, and segmentation is applied when the system predicts a segmenting punctuation sign with a ratio above threshold.
Results show to be much better than a machine translation baseline approach.
\end{abstract}

\section{Introduction}


Language is primarily a spoken medium, as every human society has a fully functioning spoken language, and until some hundred years ago, only relatively few societies had a written language, accessible to only a small class of people \cite{Aronoff:2007}.

In order to study properties of language performance, linguists can use corpora, and in order to study properties of {\em spoken} language, speech corpora are often used. A written transcript of the speech used in these corpora, aligned at the word or sentence/utterance level increases the usability of these corpora as they facilitate search and analysis of the contents. 

Manual transcription of speech corpora is a very costly process and is therefore often unavailable. Automated speech recognition (ASR) can provide a cheap, albeit imperfect, solution, by providing a rough transcript. Manual transcription can then be redefined as a post-editing process on the output of the ASR system. 
For Belgian Dutch spoken data we can use the relatively recent ASR system developed by \citeasnoun{ASR}. 

\begin{figure}
    \centering
    \includegraphics[trim={0 4cm 0 0},clip,width=12cm]{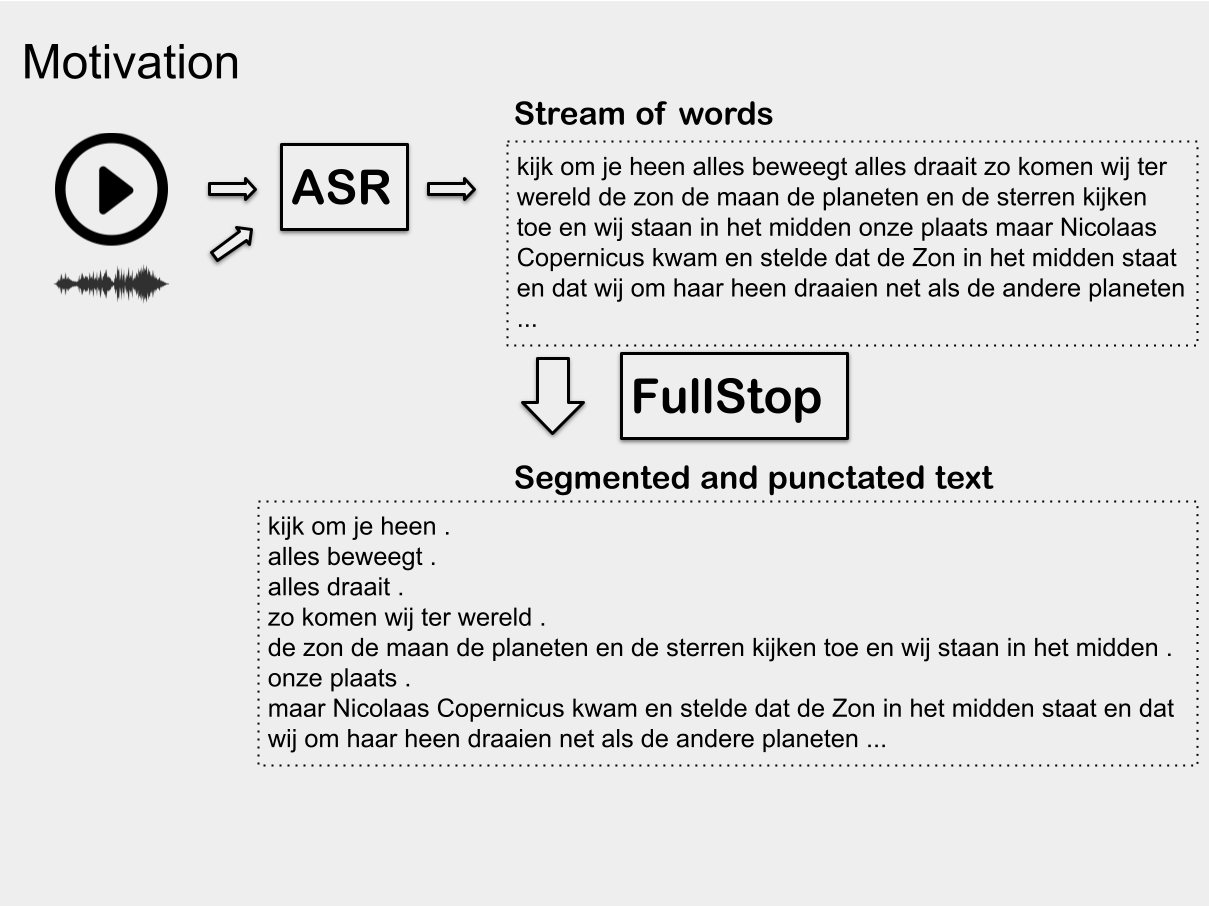}
    \caption{An incoming sound is recognized with an ASR system, resulting in a stream of words. The FullStop approach, presented in this paper, segments this stream of words in segments, by predicting punctuation.}
    \label{motivation}
\end{figure}

As shown in Figure \ref{motivation}, incoming Belgian Dutch speech is recognized by an ASR system, resulting in a transcript consisting of a stream of words with time stamps (not shown in Figure \ref{motivation}), but without any segmentation (into sentences or utterances) or punctuation. While this already makes it possible to search for the occurrence of specific words in the speech, the streams have a low readability because of the lack of segmentation.

The FullStop model for Dutch provides a punctuation and segmentation prediction system, that consists of two steps:

\begin{enumerate} 
    \item A sliding window of 200 words slides over the input text that needs to be segmented. Per window the system checks how often segmenting punctuation is predicted. If this relative frequency is above a threshold $\theta$, then the predicted punctuation is accepted and segmentation is applied as well. Increasing $\theta$ will result in a higher precision at the price of a lower recall.  We have two experimental conditions with respect to the set of segmenting punctuation $S$: only the full stop ($S =\{.\}$), and the full stop and the question mark ($S = \{.,?\}$).
    \item This step takes as input the 200 words from the previous step and segments it in batches of up to 512 tokens. This is necessary since the used transformer-based models are limited to 512 tokens input length. The model predicts for every token whether it is followed by a punctuation sign that is part of $P =$ \{:-,?.0\}, where $0$ indicates that the classifier predicts that no punctuation follows. 
\end{enumerate}

Together, these two steps apply the classifier of step 2 onto every sliding window of 200 words, implying that we get a maximum of 200 punctuation predictions per word, depending on in how many sliding windows the word appears. The ratio of predictions of a certain punctuation should be above $\theta$ before it is accepted. 

\bigskip

One type of speech corpora that particularly motivates this study is the preprocessing of {\em multimedia} corpora, consisting of video and speech.
Video is becoming an increasingly popular means of communication, with 300 hours of video being uploaded to YouTube every minute,\footnote{\url{https://fortunelords.com/youtube-statistics/}} and TikTok becoming ever more popular. 
It therefore makes sense that the creation of video corpora becomes more important. 
Audio transcription of these video corpora is often the first step, and the usage of ASR helps in the transcription process. 
There are two video corpora that motivate the current research:

\begin{enumerate}
    \item The Spoken Academic Belgian Dutch corpus, and
    \item The Belgian Federal COVID-19 Sign language (BeCoS) corpus.
\end{enumerate}

\paragraph{Spoken Academic Belgian Dutch (SABeD)}
The first corpus is the Spoken Academic Belgian Dutch (SABeD) corpus,\footnote{\url{https://www.arts.kuleuven.be/ling/language-education-society/projects/sabed}} which is currently under development. 
The SABeD project is an interdisciplinary research project which develops a corpus of spoken academic Belgian Dutch consisting of at least 200 lectures.

Lectures are typical of higher education. In lectures students learn new course content in a language register they are not familiar with, viz. academic Dutch. The SABeD project will
\begin{itemize}
    \setlength\itemsep{0.3em}
    \item compile a corpus of spoken academic Belgian Dutch;
    \item investigate the effectiveness of ASR for automatic transcription of spoken texts;
    \item create a word frequency list of spoken academic Belgian Dutch and 
    \item develop a vocabulary test of spoken academic Belgian Dutch. 
\end{itemize} 

The corpus is mainly based on video lectures of academic teaching to first-year bachelor students. These videos are, as a positive consequence of the COVID-19 pandemic, abundantly available. The lectures contain content in a language register students are not familiar with, viz. academic dutch. The corpus will serve as a source to develop representative study and test material on academic language. Furthermore, it can also serve as a source for linguistic research and a tool to optimize the language policy of higher education institutions. It will allow us to create study material and tests for international students and will be an important tool for researchers, language support and policy makers.

An additional aim is the improvement of Belgian Dutch ASR through the creation of a spoken Dutch corpus with manually transcribed (or corrected) speech. Based on these manual transcriptions the ASR system of \citeasnoun{ASR} will be retrained to improve fully automated transcriptions at a later stage so the corpus can be expanded efficiently. Once ready, the corpus will be made freely available for research through the Dutch Language Institute and the CLARIN Virtual Language Observatory.\footnote{\url{https://vlo.clarin.eu}} 

In order to speed up the transcription process, the first step consists of applying ASR. As a second step, human transcribers edit and correct the ASR output. A tool that provides manual transcription functionality is ELAN \cite{elan}. ELAN is an annotation tool for audio and video recordings and supports the creation of multiple tiers of annotation. We integrate the ASR output into ELAN by converting the automatically recognized words and associated time stamps into an ELAN tier. The human editor can easily adapt the content of the tier, but the unit of annotation in the tier is the word, as this is the unit of output of the ASR system. This is not the most convenient unit for manual transcription, due to the fact that ASR errors often lead to errors over the word boundaries. Human editors would have to make changes over several units and adapt unit boundaries (to keep the alignment with the audio/video), which is quite a time-consuming task in ELAN. A much more convenient unit to work with is the sentence or utterance level. The model described in this paper provides a way to segment the ASR output stream into appropriate segments, usable for human annotation. 

Within the SABeD corpus we are not going to provide manually corrected transcripts for the entire videos, but limit these to the first 25 minutes and the last 5 minutes, to keep the data set balanced, irrespective of the length of the videos. The rest of the video corpus will be released with fully automatic transcripts (and fully automatic segmentation). 

\paragraph{The Belgian Federal COVID-19 Sign language (BeCoS) corpus}
The second corpus that motivates the described punctuation and segmentation approach is the BeCoS corpus, the Belgian federal COVID-19 Sign language video corpus. This corpus is extensively described in \citeasnoun{covid}, and is developed within the SignON project.\footnote{\url{https://www.signon-project.eu/}}
SignON is a user-centric and community-driven project that aims to facilitate the exchange of information among Deaf, hard of hearing and hearing individuals across Europe, targeting the Irish, British, Dutch, Flemish and Spanish sign as well as the English, Irish, Dutch and Spanish spoken languages.

One of the bottlenecks for developing machine translation (MT) systems between sign languages and spoken/written languages is the lack of parallel data. The BeCoS corpus addresses this issue. It consists of 220 press conferences of the Belgian Federal Government concerning the COVID-19 pandemic, totalling 178 hours of speech. These press conferences were live interpreted into sign language: when speech was in Dutch, the sign language was {\em Vlaamse GebarenTaal} (VGT, Flemish Sign Language), when speech was in French, the sign language was {\em Langue des Signes de Belgique Francophone} (LSFB, Belgian Francophone Sign Language). 
The Dutch-VGT part of the data can serve as a parallel corpus for training a machine translation engine from VGT to Dutch or vice versa. 

The speech in this corpus is unscripted, and manual transcription is currently unattainable. As MT engines commonly are trained on parallel data at the sentence level, the corpus requires a sentence-like segmentation, which is attained by the models proposed in this paper. \\

Segmenting transcriptions generated by ASR systems has an application in voice user interfaces as well. We developed the first FullStop model to segment multi sentence user statements into single statements. The goal was to process the sentences of users' utterances individually. This is important as typical text classification models can only classify the users' intention, e.g a command, reliably if the input text does not contain multiple intentions. 

\begin{center}
    ***
\end{center}


Section \ref{related} presents related work. Section \ref{data} describes the data sets we used and how they were processed, section \ref{model} presents the models, and section \ref{experiments} first describes an experimental evaluation as a classifier, then performs a qualitative discussion on some system output, and ends with showing results on full stop prediction on out of domain data, using the sliding window on continuous text streams. The final section, section \ref{conclusions} concludes the paper.

\section{Related Work}
\label{related}
\citeasnoun{pais2021} provides an extensive survey of what they call punctuation restoration, making a distinction between methods that only use lexical features and methods that include audio specific features. As we work on the recognition output of the ASR, we do not consider the latter. 

Within the methods with lexical features, there are the early rule-based approaches and bootstrapping approaches that extract rules from large corpora, such as \citeasnoun{petasis2001}. There are also the $n$-gram approaches, such as \citeasnoun{Stolcke1996} for sentence boundary detection. Conditional random fields are used by Lu and Ng (2010) amongst others. Character-level recurrent neural networks have been used by \citeasnoun{susanto2016}. \citeasnoun{Tilk2016} approach the punctuation restoration problem as a bidirectional recurrent neural network with attention model. 

\bigskip

\citeasnoun{sepp} lists other sources of punctuation/segmentation research.  \citeasnoun{attia-etal-2014-gwu} constitutes a rather traditional approach to spelling and punctuation correction for Arabic. Classification is carried out with Support Vector Machines and Conditional Random Field (CRF) classifiers,
using part-of-speech and morphological information, and obtains an
F1-score of 0.56, with the CRF classifier and a window size of five tokens.

\citeasnoun{che2016} experiments with different neural network architectures, using pretrained GloVe embeddings \cite{glove} as inputs. It evaluates its models on ASR transcripts of TED talks, predicting commas, periods, and question marks. Its best result in this 4-class classification  is an F1 -score of 0.54.

\citeasnoun{sunkara-etal-2020-robust} works in the clinical domain on the output of medical ASR systems. It jointly models punctuation and truecasing by predicting a punctuation sequence and then the case of each input word. It uses a pretrained transformer model in combination with subword embeddings to overcome lexical sparsity in the medical domain. It carries out a fine-tuning step on medical data and a task adaptation step, randomly masking punctuation marks, before training the actual model. Predicting full stops and commas, it achieves F1 -scores of 0.81 (for commas) and 0.92 (for full stops) with Bio-BERT \cite{biobert}, which
was trained on biomedical corpora.


Previous work on multilingual punctuation prediction is described in \citeasnoun{li2020} and \citeasnoun{guerreiro2021}. 

\bigskip

\citeasnoun{vandeghinste2018} models punctuation prediction in the context of speech translation, but also investigates a monolingual approach for Dutch, modelling punctuation prediction as a machine translation problem, in which the source language is the text without punctuation, and the target language is the text with punctuation. It shows that a neural MT approach that uses LSTM cells works much better than a language modelling approach using LSTMs, scoring on in domain data for the punctuation set $P = \{.,?!:;()/-\}$ an $F_1$ of 0.82. We have used this approach, but now using a transformer model, as a baseline in the experiments in section \ref{outofdomain}. A statistical MT approach using Moses \cite{moses} is shown to work at least equally well, with $F_1 = 0.83$.


\bigskip

In 2021 there was the shared task in Sentence End and Punctuation Prediction in NLG Text (SEPP-NLG) \cite{SEPP_SharedTask},\footnote{\url{https://sites.google.com/view/sentence-segmentation}} which consisted of two subtasks:
\begin{enumerate}
    \item Fully unpunctuated sentences - full stop detection: Given the textual content of an utterance where the full stops are fully removed, correctly detect the end of sentences by placing a full stop in appropriate positions.
    \item Fully unpunctuated sentences - full punctuation marks: Given the textual content of an utterance where all punctuation marks are fully removed, correctly predict all punctuation marks. 
\end{enumerate}

\citeasnoun{sepp} modelled this task as a token-wise prediction and examined several language models based on the transformer architecture. They trained two separate models for the two tasks and submitted their results for all four languages of the shared task, reaching state-of-the-art F-scores. They advocated transfer learning for solving the task and showed that the multilingual transformer models yielded better results than monolingual models. It is this approach that is taken in the current paper, and which is applied and evaluated on Dutch. 

For the SEPP-NLG task \citeasnoun{sepp} also evaluated a CRF based model and found that this approach was outperformed by transformer based models. A GRU based model was submitted by \cite{MasielloRuiz2021ParticipationOH} for the shared task. This model scored 10 to 20 percent points lower $F_1$ scores than the best transformer based models. For these reasons we did not consider RNN based models or more classical ML approaches for this work.

\section{Data}
\label{data}
To finetune our models we experimented with two different data sets: Europarl \cite{europarl} and SoNaR \cite{SONAR}.

\paragraph{Europarl data set (EP)} contains transcribed plenary sessions of the European Parliament. 
For our models we used the Europarl v8 data, to be analogous with the other languages in the model. The text was extracted from the data downloads from OPUS \cite{opus}.

\paragraph{SoNaR data set} contains texts from different genres and domains in standard Dutch that have been written after 1954. The data was obtained from the SoNaR website.\footnote{\url{http://hdl.handle.net/10032/tm-a2-h5}} We found that SoNaR contains a number of artefacts like HTML code that can lead to issues when processing this data set.

For out of domain evaluation, as described in sections \ref{qual} and \ref{outofdomain}, we make use of the OpenSubtitles data\footnote{\url{http://www.opensubtitles.org/}} for Dutch, as made available on OPUS \cite{lison-tiedemann-2016-opensubtitles2016}. OpenSubtitles is a large database of movie and TV subtitles. We chose this data as it mainly contains translations of {\em spoken} language.

\paragraph{Data Preprocessing}
Data was split at the sentence level and tokenized at the word level using the Moses corpus preprocessing tools included in the Moses3.0 distribution \cite{moses}. All data was truecased, as the ASR output is also truecased. The data was then converted into a tab-separated format, where the first column contains the word, the second column contains a 0 if the word is not followed by a full stop and a 1 if the word is followed by a full stop (to allow for binary classification as sentence segmentation), and the third column contains a 0 if the word is not followed by punctuation but contains the punctuation sign if it is followed by it. An example is given in Table \ref{sepp}. This format is consistent with the format used in the shared task on Sentence End and Punctuation Prediction in NLG Text (SEPP-NLG 2021) held at SwissText (Swiss Text Analytics Conference) in 2021.

For both data sets we split the data into 75\% training data and 25\% test data. For the SoNaR data set we needed to downsample the training data to 1 GB, due to limited computing resources.

\begin{table}[]
    \centering
    \begin{tabular}{lll}
    \hline
    ...\\
    doos&	0&	0\\
    van	&0&	0\\
    pandora&0&	0\\
    zouden&	0&	0\\
    openen&	0&	.\\
    hoe&	0&	0\\
    ...\\
    op&	0&	0\\
    de&	0&	0\\
    volgende&	0&	0\\
    vraag&	0&	:\\
    kunnen&	0&	0\\
    ...\\
    \hline
    \end{tabular}
    \caption{Sepp format for text {\em ... doos van pandora zouden openen. hoe ... op de volgende vraag: kunnen ...}}
    \label{sepp}
\end{table}

\section{The Models}
\label{model}
Transformers \cite{vaswani} and combining transformers with transfer learning \cite{bert} have led to performance gains for many different NLP tasks. 

The first model we present here is an extension for Dutch of the models of \citeasnoun{sepp}.

We trained a token classification model, based on the Dutch language model RobBERT \cite{delobelle-etal-2020-robbert}. We also trained a Dutch model based on BERTje \cite{bertje}, but found that RobBERT slightly outperformed BERTje with a 0.75\% better $F_1$ score. 
For every token in the input sequence, the model predicts a punctuation marker that follows the token. The model is trained to predict punctuation marks of the set $P = \{:-,?.0\}$, with 0 indicating that the word is not followed by a marker. 
We finetuned several variants of this model on the two different datasets. These variants are shown in Table \ref{tab:models}.

Transformer models can only process sequences of a fixed length, typically 512 tokens. Therefore we implemented a sliding window approach to process the documents in our data set, which are typically longer than 512 tokens. 
The simplest method to achieve that is by splitting the text into chunks of 200 words before processing. The number of 200 words was chosen empirically to account for the fact that words get tokenized into more than one token (subword tokenization). With this method, it is important to leave some headroom since some words get decomposed into multiple tokens. This problem is more prominent in languages that allow compound words, such as Dutch. 

\bigskip

We choose to train a multilingual model for this work as well. A multilingual model can simplify the data processing in mixed language scenarios. In our previous work \cite{sepp} we found that multilingual models perform on par and in some cases, like Italian, significantly better than monolingual models. We wanted to investigate if this is the case for the Dutch language as well and trained a model on Dutch, English, German, French, and Italian texts.
For the multilingual model, we used ”xlm-roberta-base” \cite{conneau2020}. We previously evaluated a list of current language models and gained the best results using XLM Roberta in multilingual settings.

We choose to use the same hyper-parameters that we evaluated in our previous work. This hyper-parameters search included the optimiser algorithm, learning rate, random initialisation seed. All models were trained for 3 epochs using Adafactor \cite{pmlr-v80-shazeer18a} and a learning rate of $4e^{-5}$ with a batch size of 8 and 16 as the seed. Furthermore we used 16-bit-precision training to improve training and inference efficiency.


\bigskip

As explained in section \ref{data}, for our experiments we used two different data sets: Europarl \cite{europarl} and SoNaR \cite{SONAR}. We trained one model for each data set, as well as a multilingual model on Dutch, English, German, French, and Italian sentences from the Europarl data set. For this model we used about 400 MB of data per language. Lastly we tested a combination of all available data from the SoNaR and Europarl data set. For this model the Dutch data set was downsampled to be on par with the other languages and contains 200 MB of Europarl and 200 MB of SoNaR data. Table \ref{tab:models} provides an overview of the trained models.

\begin{table}[h]
    \centering
    \begin{tabular}{l|l|l}    
        Model Name & Data Set & Base Model \\ \hline
        Monolingual Europarl & Nl EuroParl & RobBERT \\ 
        Monolingual SoNaR & Nl SoNaR & RobBERT \\ 
        Multilingual EP & Nl, En, Fr, De, It Europarl & xlm-roberta-base \\ 
        Multilingual EP+SoNaR & Nl, En, Fr, De, It Europarl + Nl SoNaR & xlm-roberta-base \\
        \hline
    \end{tabular}
    \caption{The list of data set and model combinations that we evaluated for this work.}
    \label{tab:models}
\end{table}

The models described here are made available on HuggingFace.\footnote{\url{https://huggingface.co/oliverguhr/}} We also provide a high level software library to simplify the usage of the models.\footnote{\url{https://github.com/oliverguhr/deepmultilingualpunctuation}}

\section{Experimental Results}
\label{experiments}
In section \ref{classifier} we describe the evaluations of different variants of the model, evaluated as a multiclass classifier. Section\ref{qual} shows some qualitative evaluation. Section \ref{outofdomain} describes how the model performs on out-of-domain data.

\subsection{Evaluation as a classifier}
\label{classifier}

Table \ref{tab:modelperclassf1} compares the per class $F_1$ scores for each model on the test set of the corresponding data set it was trained on. Tables with the precision and recall values are presented in the Appendix. The overall macro and micro averaged $F_1$ scores are in the same range for every model. There are more pronounced differences for certain classes. Question marks and full stops for models including SoNaR data have 10 to 13 percentage points lower scores than models trained on Europarl without SoNaR. We assume the reason for this is that the SoNaR data set contains more diverse data and more noise (e.g. HTML code) than Europarl and is, therefore, harder to learn for the model. 

\begin{table}[h]
      \centering
      \begin{tabular}{lcccc}
      \hline
   Label/Model & EP  & SoNaR & Multilingual EP & Multilingual EP + SoNaR \\
        \hline

0 &  0.994  &  0.986 & 0.994 & 0.987 \\
. &  0.961  &  0.855 & 0.959 & 0.854 \\
, &  0.811  &  0.721 & 0.813 & 0.723 \\
? &  0.849  &  0.687 & 0.817 & 0.671 \\
- &  0.462  &  0.723 & 0.464 & 0.613 \\
: &  0.655  &  0.697 & 0.657 & 0.709 \\
macro $F_1$ & 0.789 &  0.778 & 0.784 & 0.760 \\
micro $F_1$ & 0.983 &  0.964 & 0.983 & 0.965 \\
        \hline
      \end{tabular}
      \caption{\label{tab:modelperclassf1}
      Per class $F_1$ scores for the FullStop models on the Dutch test data sets. For detailed evaluation results of each model and per class precision and recall metrics please see the appendix. }
\end{table}

Table \ref{tab:modelperclassf1} also shows that the multilingual Europarl model improved the $F_1$ scores in every class over its monolingual version. Part of this improvements can be explained by the fact that XLM-Roberta base uses more than twice as many parameters than RobBERT. 
Note that it is not possible to compare the performance of the other models directly, since models using the SoNaR data set in Table \ref{tab:modelperclassf1} where trained and tested on different data sets. Therefore we compared the models using an out of domain data set in section \ref{outofdomain}.

\begin{figure*}[t]
     \centering
     \begin{subfigure}[b]{0.49\textwidth}
         \centering
         \caption{Monolingual Europarl}
         \includegraphics[trim={0 5 0 40},clip,width=\textwidth]{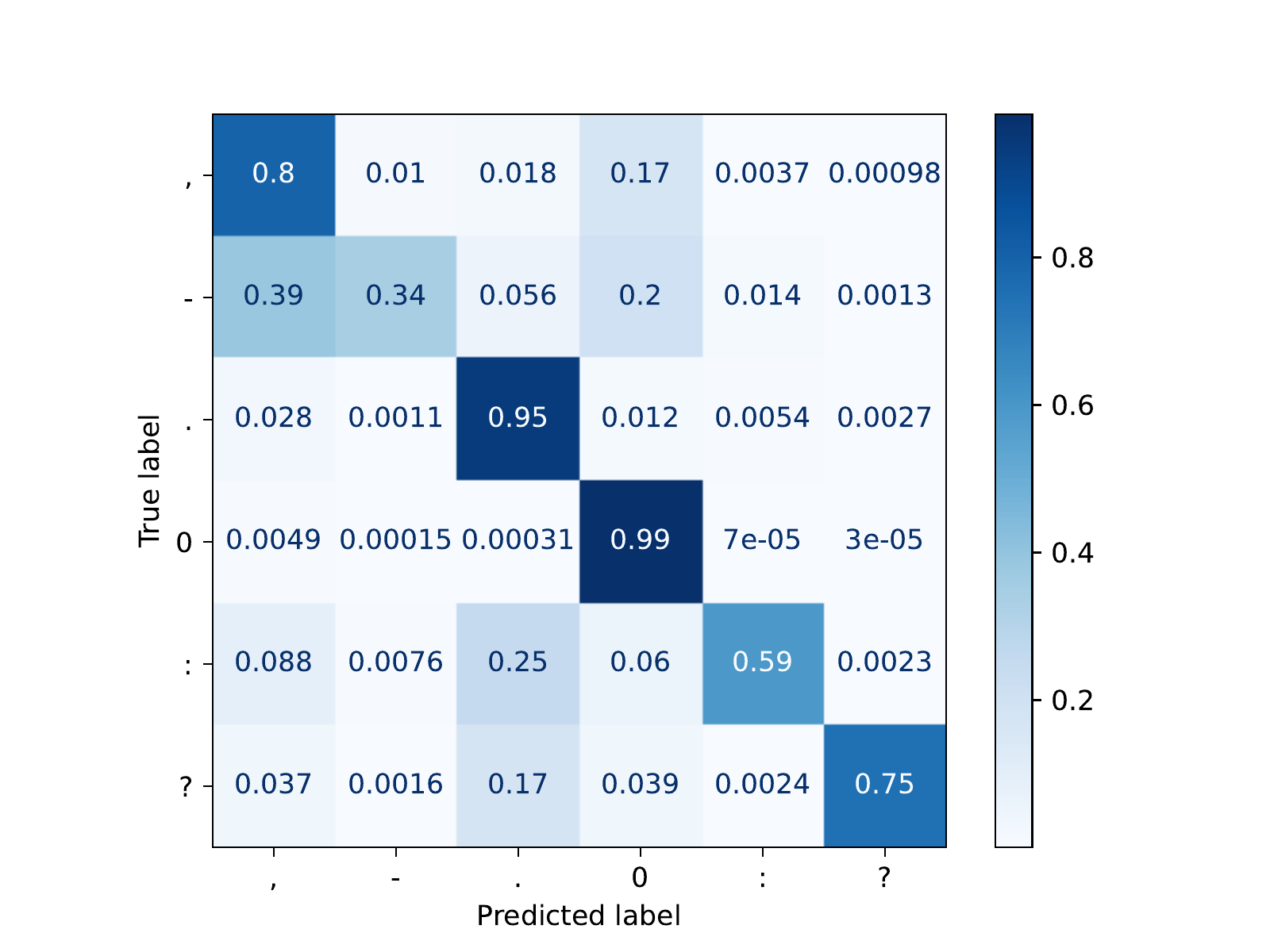}
         \label{fig_}
     \end{subfigure}
     \hfill
     \begin{subfigure}[b]{0.49\textwidth}
         \centering
         \caption{Monolingual SoNaR}
         \includegraphics[trim={0 5 0 40},clip,width=\textwidth]{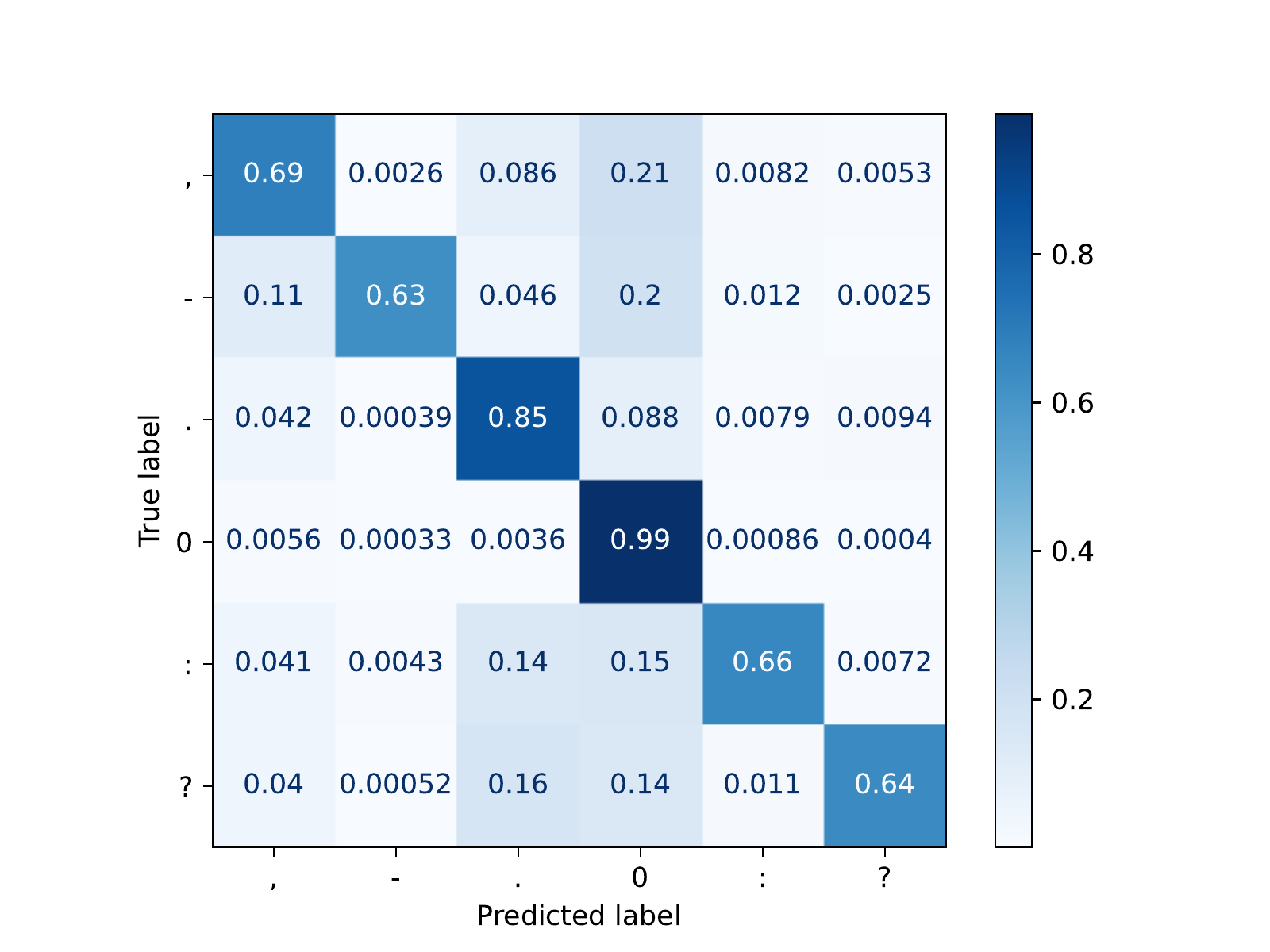}
         \label{fig_bert_balanced}
     \end{subfigure}
     
     \begin{subfigure}[b]{0.49\textwidth}
         \centering
         \caption{Multilingual EP}
         \includegraphics[trim={0 5 0 40},clip,width=\textwidth]{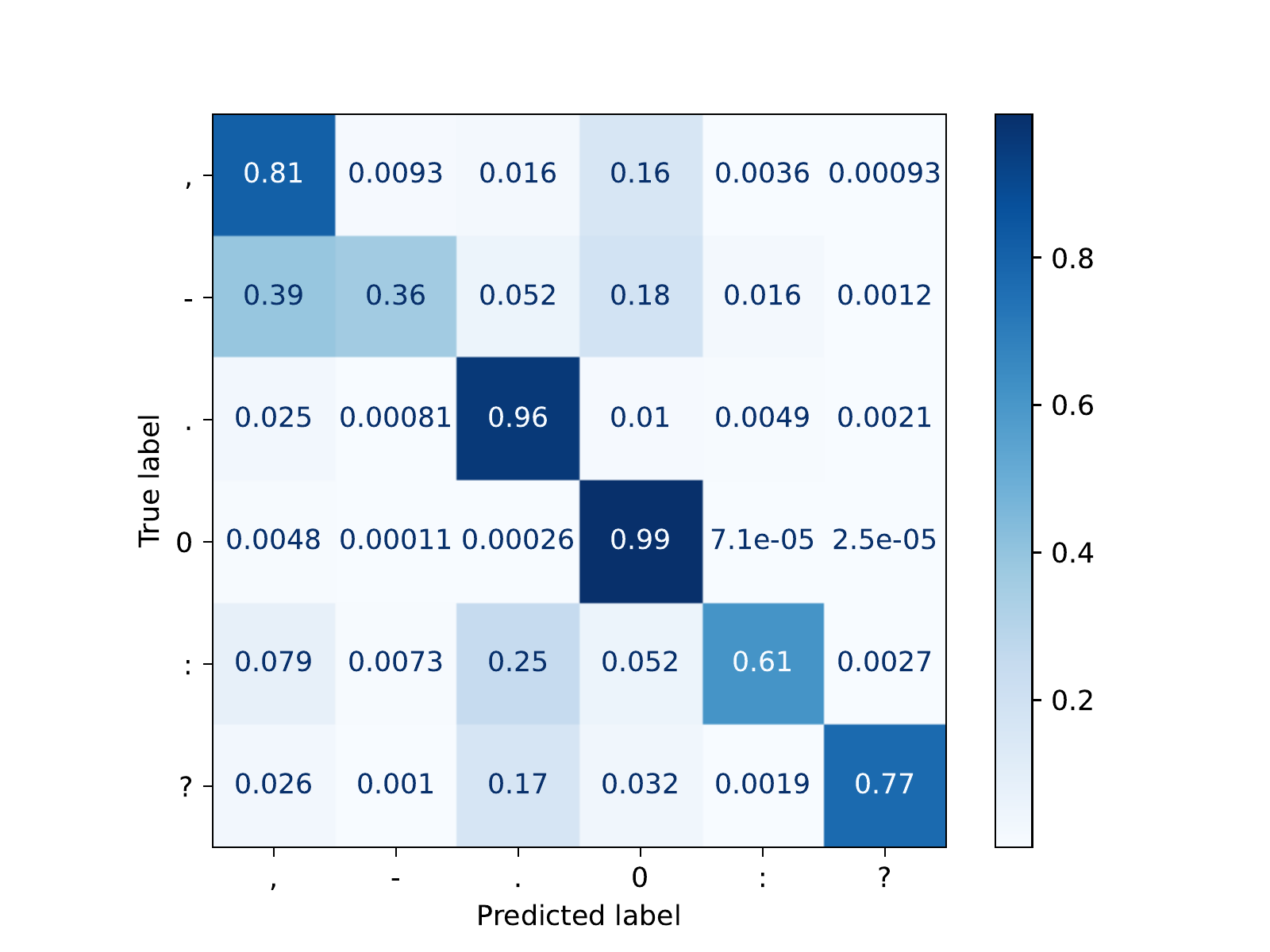}
         \label{fig_ft_unbalanced}
     \end{subfigure}
    \hfill
    \begin{subfigure}[b]{0.49\textwidth}
         \centering
         \caption{Multilingual EP+SoNaR}
         \includegraphics[trim={0 5 0 40},clip,width=\textwidth]{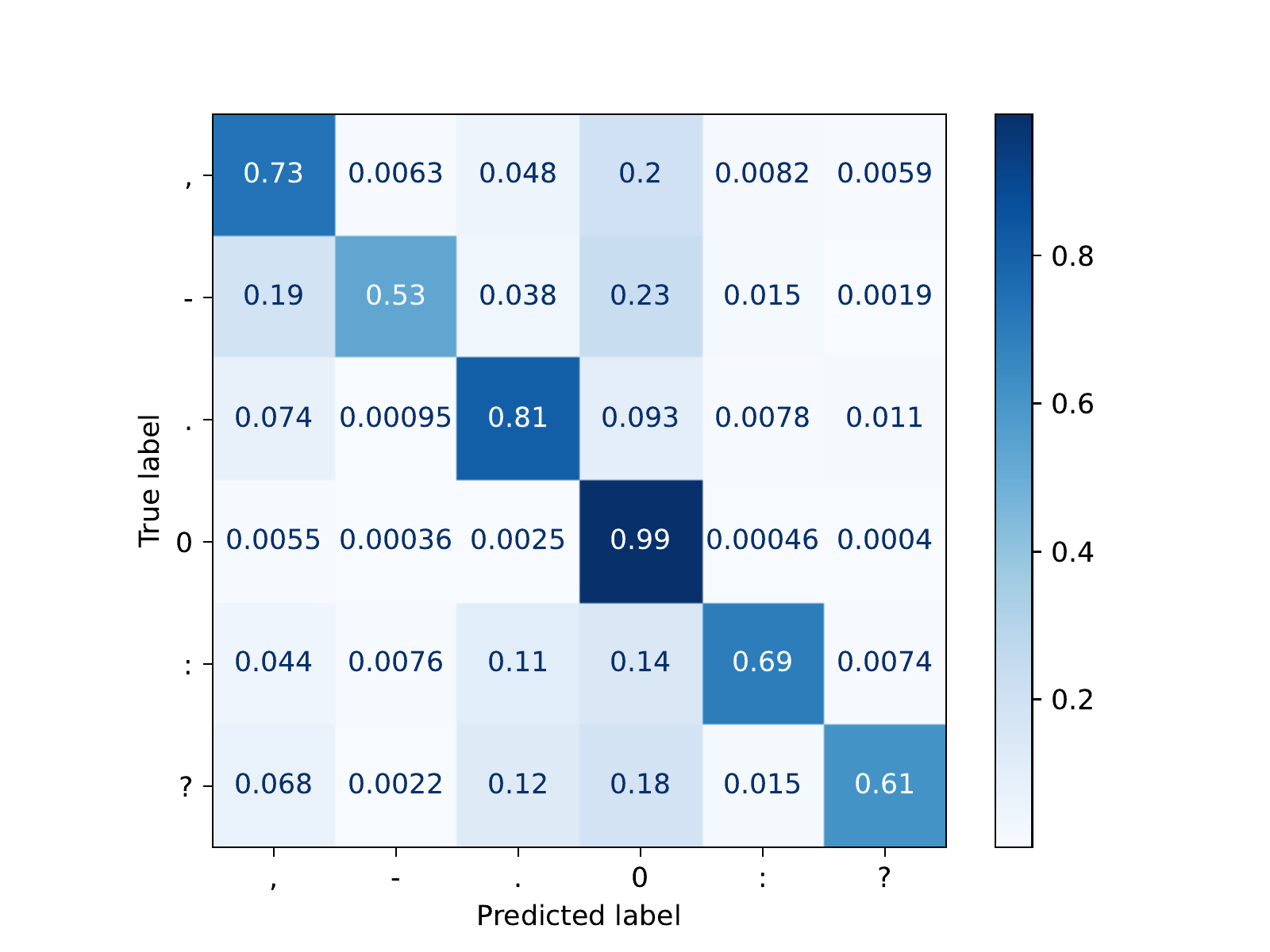}
         \label{fig_bert_unbalanced}
     \end{subfigure}
        \caption{Confusion matrices for Dutch language for the FullStop models. Note that all values are rounded.}
        \label{fig:cm_nl}
\end{figure*}

Figure \ref{fig:cm_nl} shows a confusion matrix for the Dutch language for every trained model. Models trained on Europarl tend to confuse dashes with commas. SoNaR based models predict 14\% to 18\% of the colons and question marks as 0 or no punctuation mark. All models predict 10\% to 25\% of the colons and question marks with full stops. This is to be expected, as colons are more stylistic markers and there are no strict usage rules. Overall the Dutch language results are in line with English, French, German and Italian language predictions from our previous work. 


Furthermore we compared the performance of the different languages that both multilingual model where trained on, in Tables \ref{tab:multilingual_europarl} and \ref{tab:multilingual_europarl_sonar}. We think these models are useful in scenarios where users mix languages or the source language is unknown, for example in social media posts. The evaluation results of the multilingual Europarl model (Table \ref{tab:multilingual_europarl}) are comparable between all five languages. We see an overall drop in Dutch language performance for the multilingual model using both Europarl and SoNaR data in Table \ref{tab:multilingual_europarl_sonar}. This is to be expected, as the SoNaR data set is more diverse. The results of the other four languages remain the same for both models.

For efficiency reasons we choose to train XLM-Roberta base instead of large models. Comparing the results from Tables \ref{tab:multilingual_europarl} and \ref{tab:multilingual_europarl_sonar} with the finetuned multilingual model from our previous work, we estimate that a size "large" model could improve the macro $F_1$ by 5\%. However, XLM-Roberta large models use more than twice as many parameters as base models, with 550 million parameters compared to 270 million.

\begin{table}[h]
      \centering
      \begin{tabular}{lccccc}
      \hline
   Label & EN  & DE & FR & IT & NL \\
        \hline
          0 & 0.990 &  0.996 & 0.991 & 0.988 & 0.994 \\
          . & 0.924 &  0.951 & 0.921 & 0.917 & 0.959 \\
          , & 0.798 &  0.937 & 0.811 & 0.778 & 0.813 \\
          ? & 0.825 &  0.829 & 0.800 & 0.736 & 0.817 \\
          - & 0.345 &  0.384 & 0.353 & 0.344 & 0.464 \\
          : & 0.535 &  0.608 & 0.578 & 0.544 & 0.657 \\
macro $F_1$ & 0.736 &  0.784 & 0.742 & 0.718 & 0.784 \\
micro $F_1$ & 0.975 &  0.987 & 0.977 & 0.972 & 0.983 \\

        \hline
      \end{tabular}
      \caption{\label{tab:multilingual_europarl}
       Per class $F_1$ scores of the multilingual Europarl model. Tested on English, German, French, Italian and Dutch language on the test data set. }
\end{table}

\begin{table}[h]
      \centering
      \begin{tabular}{lccccc}
      \hline
Label & EN  & DE & FR & IT & NL \\
\hline

          0 & 0.990 &  0.996 & 0.991 & 0.988 & 0.987 \\
          . & 0.924 &  0.950 & 0.921 & 0.917 & 0.854 \\
          , & 0.797 &  0.937 & 0.810 & 0.778 & 0.723 \\
          ? & 0.823 &  0.826 & 0.802 & 0.731 & 0.671 \\
          - & 0.349 &  0.380 & 0.359 & 0.348 & 0.613 \\
          : & 0.533 &  0.606 & 0.576 & 0.541 & 0.709 \\
macro $F_1$ & 0.736 &  0.783 & 0.743 & 0.717 & 0.760 \\
micro $F_1$ & 0.975 &  0.987 & 0.977 & 0.972 & 0.965 \\

        \hline
      \end{tabular}
      \caption{\label{tab:multilingual_europarl_sonar}
       Per class $F_1$ scores of the multilingual Europarl + SoNaR model. Tested on English, German, French, Italian and Dutch language data on the test set. }
\end{table}

\subsection{Qualitative Evaluation}
\label{qual}
To better understand the capabilities and the limitations of the model, we qualitatively discuss some examples, presented in Table \ref{tab:qualitative_evaluation}. The examples are selected from the OpenSubtitles corpus \cite{lison-tiedemann-2016-opensubtitles2016}.

The results that are shown are those of a single input of the words to the classifier, so there is no effect of $\theta$ in Table \ref{tab:qualitative_evaluation}.

We can see that the {\em Gold} strings of the examples contain different punctuation signs, belonging to the set $P=\{.,?\}$.They are tokenized at the word level and truecased. The {\em Input} strings simulate the stream of words as coming from an ASR system, without any punctuation. The {\em Prediction} strings show the output of the classification model, not in SEPP format, but in string format. In the examples we have marked the inserted punctuation in the Prediction in \colorbox{green}{green} where they are correct and in \colorbox{red}{red} when there is a mismatch (be it a deletion, substitution or insertion) between the Prediction and the Gold version. For ease of reference we have indexed the predicted or omitted punctuation with a subscript.

The first example shows the stream of words that was also presented in Figure \ref{motivation}, but now somewhat longer. Prediction 1 is a comma where the gold standard has a full stop. This is counted as a substitution error, but it is not an implausible prediction and could be correct. The next 5 predictions, containing comas and full stops, are correct. Then our model misses a full stop at prediction 7, which may be explained by the next sentence starting with the conjunction {\em en}. Then two more correct predictions and an omission of a comma (10), which can be attributed to the next word being {\em en} again. As Dutch has no Oxford comma rule, you would not necessarily expect a comma here. In (11) the system substitutes a full stop with a comma. A comma would be plausible at this position. (12) is a correct full stop prediction. In (13) and (14) question marks should have been predicted, but there is nothing in the word order that indicates that these are questions, so this information would have to come from the intonation, which is not available to our model. Then the system misses the final full stop (15). This may be due to the lack of context.

The second example starts with an insertion (1) of a comma. This could be considered correct if we would put the apposition {\em hier} between commas, as is done often. The next comma, ending the apposition is correctly predicted. (3) and (4) omit commas at the start of a relative phrase, which is not uncommon. (5), (6) and (7) are correctly predicted full stops and commas. (8) is an omitted comma before {\em en}. (9) is a correctly predicted full stop before an {\em en}. (10) omits a comma before a subordinate clause. (11) correctly predicts the final full stop.

From these samples we can see that the system makes several {\em real} mistakes, but that most of the differences between the {\em Gold} version and the {\em Predicted} version can be attributed to the fact that there are often are no strict punctuation rules, and that the output of the system could be argumented for. It would be interesting to see how humans would add punctuation purely based on the input text and what the inter-annotator agreement would be, but such an exercise is outside the scope of this paper.


\begin{table}[h]
      \centering
      \begin{tabular}{ll}
        \textbf{Example 1} &  \\
        \hline
          Gold& \begin{small} \makecell{kijk om je heen . alles beweegt , alles draait . zo komen wij ter wereld . de zon ,\\
          de maan , de planeten en de sterren kijken toe . en wij staan in het midden .\\
          onze plaats . maar Nicolaas Copernicus kwam en stelde dat de Zon in het midden\\
          staat , en dat wij om haar heen draaien . net als de andere planeten . een Aarde \\
          die beweegt ? maar daar zien en voelen we toch niets van ? dat was 1543 .}\end{small} \\
          Input & \begin{small}\makecell{kijk om je heen alles beweegt alles draait zo komen wij ter wereld de zon \\
          de maan de planeten en de sterren kijken toe en wij staan in het midden\\
          onze plaats maar Nicolaas Copernicus kwam en stelde dat de Zon in het midden\\
          staat en dat wij om haar heen draaien net als de andere planeten een Aarde\\
          die beweegt maar daar zien en voelen we toch niets van dat was 1543}\end{small} \\     
          Prediction & 
          \begin{small}
            \makecell{kijk om je heen \colorbox{red}{,$_1$} alles beweegt \colorbox{green}{,$_2$} alles draait \colorbox{green}{.$_3$} zo komen wij ter wereld \colorbox{green}{.$_4$} de zon \colorbox{green}{,$_5$}\\
            de maan \colorbox{green}{,$_6$} de planeten en de sterren kijken toe \colorbox{red}{ $_7$} en wij staan in het midden \colorbox{green}{.$_8$}\\
            onze plaats \colorbox{green}{.$_9$} maar Nicolaas Copernicus kwam en stelde dat de Zon in het midden\\
            staat \colorbox{red}{ $_{10}$} en dat wij om haar heen draaien \colorbox{red}{,$_{11}$} net als de andere planeten \colorbox{green}{.$_{12}$} een Aarde\\
            die beweegt \colorbox{red}{,$_{13}$} maar daar zien en voelen we toch niets van \colorbox{red}{.$_{14}$} dat was 1543 \colorbox{red}{ $_{15}$}
          }
          \end{small} \\
        \hline
        \textbf{Example 2} &  \\
        \hline
          Gold& \begin{small}\makecell{en waar we nu zitten hier , dat is bij een fotografische kijker , die heel veel gelijkenis\\
          vertoont met de kijker , die werd gebruikt door David Gill . zo ' n kijker moet dus in\\
          staat zijn om foto ' s te nemen . maar als je foto ' s neemt met die kijker , moet je \\
          natuurlijk ook een oogje houden op het stukje hemel waar hij op gericht is , en zorgen \\
          dat de kijker heel nauwkeurig de dagelijkse beweging van de hemel volgt . en daarom \\
          is zo ' n kijker zo gebouwd , dat hij een gedeelte heeft waar de fotografische plaat zich \\
          bevindt .} \end{small}\\
          Input & \begin{small} \makecell{en waar we nu zitten hier dat is bij een fotografische kijker die heel veel gelijkenis\\
          vertoont met de kijker die werd gebruikt door David Gill zo ' n kijker moet dus in\\
          staat zijn om foto ' s te nemen maar als je foto ' s neemt met die kijker moet je\\
          natuurlijk ook een oogje houden op het stukje hemel waar hij op gericht is en zorgen\\
          dat de kijker heel nauwkeurig de dagelijkse beweging van de hemel volgt en daarom\\
          is zo ' n kijker zo gebouwd dat hij een gedeelte heeft waar de fotografische plaat zich\\
          bevindt} \end{small} \\     
          Prediction & \begin{small} \makecell{en waar we nu zitten \colorbox{red}{,$_1$} hier \colorbox{green}{,$_2$} dat is bij een fotografische kijker \colorbox{red}{ $_3$} die heel veel gelijkenis\\
          vertoont met de kijker \colorbox{red}{ $_4$} die werd gebruikt door David Gill \colorbox{green}{.$_5$} zo ' n kijker moet dus in\\
          staat zijn om foto ' s te nemen \colorbox{green}{.$_6$} maar als je foto ' s neemt met die kijker \colorbox{green}{,$_7$} moet je\\
          natuurlijk ook een oogje houden op het stukje hemel waar hij op gericht is \colorbox{red}{ $_8$} en zorgen\\
          dat de kijker heel nauwkeurig de dagelijkse beweging van de hemel volgt \colorbox{green}{.$_9$} en daarom\\
          is zo ' n kijker zo gebouwd \colorbox{red}{ $_{10}$} dat hij een gedeelte heeft waar de fotografische plaat zich\\
          bevindt \colorbox{green}{.$_{11}$}} \end{small} \\
          \hline
      \end{tabular}
      \caption{\label{tab:qualitative_evaluation}
       We generated these examples with the FullStop SoNaR model. }
\end{table}

\subsection{Experiments on full stop prediction on out of domain data}
\label{outofdomain}
In this section we describe an evaluation on out of domain test data, i.e. test data coming from OpenSubtitles, as described in section \ref{data}.

We test segmentation in two variants: with segementation set $S=\{.\}$ and with segementation set $S=\{.?\}$.


\paragraph{Baseline}
As a baseline we tested a machine translation approach, similar to Vandeghinste et al. (2018), in which we consider texts with all punctuation and segmentation removed as the source language and the punctuated version as the target language.

We trained an OpenNMT \cite{opennmt} transformer model on a randomly resegmented version of the SoNaR corpus complemented with the Corpus Spoken Dutch \cite{oostdijk-etal-2002-experiences}. Resegmenation was random, but distributed normally with an average of 14 tokens and standard deviation of 3 tokens. These values are based on the average length and stdev of sentences in De Standaard, according to \citeasnoun{Vandeghinste_Bulte_2019}. In order to create the training data for the MT system, we removed all punctuation in the source side, and kept the original punctuation in the target side. 

In our best model and parameter setting, such an approach reached a segmentation prediction $F_1$ score of 42\%, which seems not good enough for practical usage. The low $F_1$ score is mostly due to low recall, as shown in table \ref{tab:eval_ood}.

\paragraph{Evaluation Procedure}
Evaluation was performed on 1000 sentences from OpenSubtitles, and applied with a sliding window of size 200. For every sequence of 200 words we checked where the system inserted an element of $S$. We accept a segmentation if the element of $S$ is predicted in more than $\theta$ of all cases. We tried different $\theta$ values, but a $\theta=0.1$ setting seemed to provide the best results.


\bigskip

\paragraph{Results} Table \ref{tab:eval_ood} shows the results for the different models we trained. It is clear that the current models present a major improvement over the baseline MT apprach. The best $F_1$-score for this evaluation seems to be the monolingual model trained on SoNaR only. This may be due to the fact that SoNaR contains different registers, amongst which some more colloquial forms of Dutch, that may be closer to the subtitles register than the data from Europarl. Note that the multilingual models also have good scores and an even higher precision than the monolingual models. There is also a clear improvement when using $S=\{.?\}$ as set of segmenters over just using $S=\{.\}$. 


\begin{table}[h]
    \centering
    \begin{tabular}{r|crrr}
        Model&$S$&Precision&Recall&$F_1$-score\\
        \hline
         Baseline ONMT&$\{.\}$&0.7187&0.2986&0.4219\\
         FullStop EP&$\{.\}$&0.8939&0.7609&0.8221\\
         FullStop SoNaR&$\{.\}$&0.8734&0.8750&0.8742\\
         FullStop Multilingual&$\{.\}$&0.9010&0.7719&0.8314\\
         FullStop Multilingual EP+SoNaR&$\{.\}$&0.9021&0.7915&0.8424\\
         \hline
         FullStop EP&$\{.?\}$&0.8962&0.8193&0.8561\\
         FullStop SoNaR&$\{.?\}$&0.8749&{\bf 0.9380}&{\bf 0.9053}\\
         FullStop Multilingual&$\{.?\}$&0.9013&0.8330&0.8658\\
         FullStop Multilingal EP+Sonar&$\{.?\}$&{\bf 0.9040}&0.8504&0.8764\\
         \hline
    \end{tabular}
    \caption{Evaluation results on out of domain data}
    \label{tab:eval_ood}
\end{table}
At this point, we are interested in the significance levels: do the models differ significantly or not, and what is the 95\% confidence interval?

\begin{table}[]
    \centering
    \begin{tabular}{llll|rrr|rr}
        \multicolumn{4}{l}{Condition}&\multicolumn{3}{l}{Characteristics}&\multicolumn{2}{l}{95\% Conf.}\\
         &Model&$\theta$&$S$&Median&Average&Std dev&Lo&Hi\\
         \hline
         A&SoNaR&0.1&$\{.\}$&0.7975&0.7785&0.0818&0.5918&0.8567\\
         B&SoNaR&0.2&$\{.\}$&0.7933&0.7746&0.0820&0.5882&0.8540\\    
         C&SoNaR&0.3&$\{.\}$&0.7892&0.7705&0.0821&0.5841&0.8505\\
         \hline
         D&SoNaR&0.1&$\{.?\}$&0.8812&0.8611&0.0839&0.6802&0.9308\\
         \hline
    \end{tabular}
    \caption{Characteristics of the distribution of $F_1$ scores over 10000 test files}
    \label{tab:characteristics}
\end{table}

\paragraph{Multiple testfiles}  
In order to determine whether the F-scores between the $S=\{.?\}$ condition and the $S=\{.\}$ condition, or the scores for different $\theta$ values differ significantly, we have tested the SoNaR model, which was the best scoring model in Table \ref{tab:eval_ood} on 10000 test sets of each 1000 sentences. These test sets were created by splitting up the OpenSubtitle file into sections of 1000 lines each. 

Per condition, we have evaluated these 10000 test sets, ranked the $F_1$ scores and taken the score at rank 251 and at rank 9750 as the values of the 95\% confidence interval.

Table \ref{tab:characteristics} lists the main characteristics of the $F_1$ scores for the different conditions that were tested on the 10000 evaluation files.

The difference between Condition A and Condition D has a p-value of 0.0981, so it is significant at the $p<.10$ level. Difference between Condition B and D is not significant ($p=0.100801$). Difference between C and D has a $p=0.091266$.

The effect of the $\theta$ parameter is not significant, but the effect of adding the question mark to $S$ is mildly significant at the $p<.10$ level.






\begin{figure}
    \centering
    \includegraphics[width=10 cm]{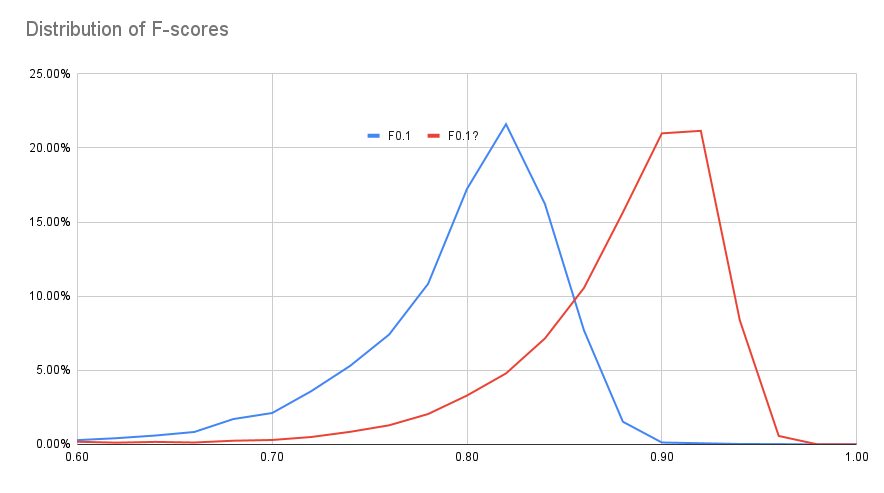}
    \caption{Distribution of $F_1$ scores over 10000 test files. The blue curve shows the distribution of using $\{.\}$ as a segmentation marker. The red curve shows the distribution when using $\{.?\}$ as segmentation markers. }
    \label{fig:distribution}
\end{figure}


Figure \ref{fig:distribution} presents a visualisation of the distribution of the $F_1$ scores for the two different $S$ conditions, which shows clearly that $S=\{.?\}$ scores better than only $S=\{.\}$.

\section{Conclusions}
\label{conclusions}
We have presented several models that perform punctuation prediction and evaluated them in different settings. We have made various models specifically for Dutch, but have also extended the multilingual model from \citeasnoun{sepp} with Dutch. The models use transfer learning from large pretrained models and are finetuned as per token classifiers. 

The models are evaluated as classifiers, reaching a similar accuracy as for the other languages, and they have also been tested on out-of-domain data, on which they present a great improvement over a baseline MT model. 


The best models are publicly available through Huggingface.\footnote{\url{https://huggingface.co/oliverguhr}}. The use of the model to actually predict segmentation on a stream of text, as used in the out-of-domain evaluation is made available on Github.\footnote{\url{https://github.com/VincentCCL/Segment_FullStop}}

For future work, it would make sense to predict different sets of tokens. As for now, we have taken the set of tokens as defined in the shared task. Training a classifier just for the prediction of segmentation, punctuation signs in set $S=\{.?!\}$ or for the prediction of all non-alphanumeric characters would be possible. The model could also be extended to predict the segmentation tokens and the true case of every word in the sequence. This is a common use case in processing the output of automatic speech recognition systems.

Another line of work would be to make a lighter version of the model, with fewer parameters, through knowledge distillation which is quicker in inference.

All in all, we can conclude that the model we present provides a usable and practical way of inserting punctuation into streams of words, and therefore turning a sequence of words into a text. The SoNaR model came out as the best model and will therefore serve this purpose in further processing of the SABeD corpus and has been used in processing of the BeCoS corpus.


\section{Acknowledgements}
Work in this paper is partly financed by the SignON project.\footnote{\url{https://signon-project.eu}} This project has received funding from the European Union's Horizon 2020 Research 
and Innovation Programme under Grant Agreement No. 101017255.
The SABeD project is funded by KU Leuven Internal Funding, Research Project 3H200610.

Oliver Guhr has been funded by the European Social Fund (ESF), SAB grant number 100339497 and the European Regional Development Funds (ERDF) (ERDF-100346119).

\bibliographystyle{clin} 
\bibliography{punctuationbib}

@inproceedings{elan,
    title = "{ELAN}: a Professional Framework for Multimodality Research",
    author = "Wittenburg, Peter  and
      Brugman, Hennie  and
      Russel, Albert  and
      Klassmann, Alex  and
      Sloetjes, Han",
    booktitle = "Proceedings of the Fifth International Conference on Language Resources and Evaluation ({LREC}{'}06)",
    month = may,
    year = "2006",
    address = "Genoa, Italy",
    publisher = "European Language Resources Association (ELRA)",
    url = "http://www.lrec-conf.org/proceedings/lrec2006/pdf/\_pdf.pdf",
}

@article{covid,
author = "Vincent Vandeghinste and Bob Van Dyck and Mathieu De Coster and Maud Goddefroy",
title = {{BeCoS corpus: Belgian Covid-19 Sign language corpus. A corpus for training Sign Language Recognition and Translation}},
year = "2022",
journal={Computational Linguistics in the Netherlands Journal},
volume={12}
}

@inproceedings{SEPP,
title={{FullStop: Multilingual Deep Models for Punctuation Prediction}},
author={Oliver Guhr and Anne-Kathrin Schumann and Frank Bahrmann and Hans-Joachim B\"ohme},
booktitle = {Swiss Text Analytics Conference. Shared task on Sentence End and Punctuation Prediction in NLG Text},
year={2021}
}

@inproceedings{MasielloRuiz2021ParticipationOH,
  title={Participation of HULAT-UC3M in SEPP-NLG 2021 shared task (short paper)},
  author={Jose Manuel Masiello-Ruiz and Jos{\'e} Luis L{\'o}pez Cuadrado and Paloma Mart{\'i}nez},
  booktitle={Swiss Text Analytics Conference. Shared task on Sentence End and Punctuation Prediction in NLG Text},
  year={2021}
}

@article{ASR, 
title={{A Hybrid ASR System for Southern Dutch}}, 
volume={11}, 
url={https://clinjournal.org/clinj/article/view/119},  
journal={Computational Linguistics in the Netherlands Journal}, 
author={Van Dyck, Bob and BabaAli, Bagher and Van Compernolle, Dirk}, 
year={2021}, month={Dec.}, 
pages={27–34} }

@inproceedings{delobelle-etal-2020-robbert,
    title = "{R}ob{BERT}: a {D}utch {R}o{BERT}a-based {L}anguage {M}odel",
    author = "Delobelle, Pieter  and
      Winters, Thomas  and
      Berendt, Bettina",
    booktitle = "Findings of the Association for Computational Linguistics: EMNLP 2020",
    month = nov,
    year = "2020",
    address = "Online",
    publisher = "Association for Computational Linguistics",
    url = "https://aclanthology.org/2020.findings-emnlp.292",
    doi = "10.18653/v1/2020.findings-emnlp.292",
    pages = "3255--3265",
}

@article{pais2021,
  title={Capitalization and punctuation restoration: a survey},
  author={P{\u{a}}i{\c{s}}, Vasile and Tufi{\c{s}}, Dan},
  journal={Artificial Intelligence Review},
  volume={55},
  number={3},
  pages={1681--1722},
  year={2022},
  publisher={Springer}
}

@inproceedings{li2020,
  title={A 43 Language Multilingual Punctuation Prediction Neural Network Model.},
  author={Li, Xinxing and Lin, Edward},
  booktitle={INTERSPEECH},
  pages={1067--1071},
  year={2020}
}

@article{guerreiro2021,
  title={Towards better subtitles: A multilingual approach for punctuation restoration of speech transcripts},
  author={Guerreiro, Nuno Miguel and Rei, Ricardo and Batista, Fernando},
  journal={Expert Systems with Applications},
  volume={186},
  pages={115740},
  year={2021},
  publisher={Elsevier}
}

@inproceedings{vandeghinste2018,
  title={A comparison of different punctuation prediction approaches in a translation context},
  author={Vandeghinste, Vincent and Verwimp, Lyan and Pelemans, Joris and Wambacq, Patrick},
  booktitle={Proceedings of the 21st Annual Conference of the European Association for Machine Translation: 28-30 May 2018, Universitat d'Alacant, Alacant, Spain},
  pages={269--278},
  year={2018},
  organization={European Association for Machine Translation}
}

@inproceedings{petasis2001,
    title = "Using Machine Learning to Maintain Rule-based Named-Entity Recognition and Classification Systems",
    author = "Petasis, Georgios  and
      Vichot, Frantz  and
      Wolinski, Francis  and
      Paliouras, Georgios  and
      Karkaletsis, Vangelis  and
      Spyropoulos, Constantine D.",
    booktitle = "Proceedings of the 39th Annual Meeting of the Association for Computational Linguistics",
    month = jul,
    year = "2001",
    address = "Toulouse, France",
    publisher = "Association for Computational Linguistics",
    url = "https://aclanthology.org/P01-1055",
    doi = "10.3115/1073012.1073067",
    pages = "426--433",
}

@INPROCEEDINGS{Stolcke1996,  author={Stolcke, A. and Shriberg, E.},  booktitle={Proceeding of Fourth International Conference on Spoken Language Processing. ICSLP '96},   title={Automatic linguistic segmentation of conversational speech},   year={1996},  volume={2},  number={},  pages={1005-1008 vol.2},  doi={10.1109/ICSLP.1996.607773   }}

@inproceedings{Tilk2016,
  title={Bidirectional Recurrent Neural Network with Attention Mechanism for Punctuation Restoration},
  author={Ottokar Tilk and Tanel Alum{\"a}e},
  booktitle={INTERSPEECH},
  year={2016}
}

@inproceedings{che2016,
    title = "Punctuation Prediction for Unsegmented Transcript Based on Word Vector",
    author = "Che, Xiaoyin  and
      Wang, Cheng  and
      Yang, Haojin  and
      Meinel, Christoph",
    booktitle = "Proceedings of the Tenth International Conference on Language Resources and Evaluation ({LREC}'16)",
    month = may,
    year = "2016",
    address = "Portoro{\v{z}}, Slovenia",
    publisher = "European Language Resources Association (ELRA)",
    url = "https://aclanthology.org/L16-1103",
    pages = "654--658"
    }

@inproceedings{susanto2016,
    title = "Learning to Capitalize with Character-Level Recurrent Neural Networks: An Empirical Study",
    author = "Susanto, Raymond Hendy  and
      Chieu, Hai Leong  and
      Lu, Wei",
    booktitle = "Proceedings of the 2016 Conference on Empirical Methods in Natural Language Processing",
    month = nov,
    year = "2016",
    address = "Austin, Texas",
    publisher = "Association for Computational Linguistics",
    url = "https://aclanthology.org/D16-1225",
    doi = "10.18653/v1/D16-1225",
    pages = "2090--2095",
}

@inproceedings{europarl,
    title = "{E}uroparl: A Parallel Corpus for Statistical Machine Translation",
    author = "Koehn, Philipp",
    booktitle = "Proceedings of Machine Translation Summit X: Papers",
    month = sep # " 13-15",
    year = "2005",
    address = "Phuket, Thailand",
    url = "https://aclanthology.org/2005.mtsummit-papers.11",
    pages = "79--86",
}

@inproceedings{opus,
    title = "Parallel Data, Tools and Interfaces in {OPUS}",
    author = {Tiedemann, J{\"o}rg},
    booktitle = "Proceedings of the Eighth International Conference on Language Resources and Evaluation ({LREC}'12)",
    month = may,
    year = "2012",
    address = "Istanbul, Turkey",
    publisher = "European Language Resources Association (ELRA)",
    url = "http://www.lrec-conf.org/proceedings/lrec2012/pdf/463\_Paper.pdf",
    pages = "2214--2218",
}

@incollection{SONAR,
    author = "Oostdijk, N. and Reynaert, M. and Hoste, V. and Schuurman, I.",
    year = 2013,
    title = {"The Construction of a 500 Million Word Reference Corpus of Contemporary Written Dutch"},
    booktitle = "Essential Speech and Language Technology for Dutch: Results by the STEVIN-programme",
    editors = "P. Spyns and J. Odijk",
    publisher = "Springer Verlag"
    }

@misc{bertje,
  doi = {10.48550/ARXIV.1912.09582 },
  url = {https://arxiv.org/abs/1912.09582},
  author = {de Vries, Wietse and van Cranenburgh, Andreas and Bisazza, Arianna and Caselli, Tommaso and van Noord, Gertjan and Nissim, Malvina},
  title = {BERTje: A Dutch BERT Model},
  publisher = {arXiv},
  year = {2019},
  copyright = {Creative Commons Attribution 4.0 International}
}

@inproceedings{conneau2020,
    title = "Unsupervised Cross-lingual Representation Learning at Scale",
    author = "Conneau, Alexis  and
      Khandelwal, Kartikay  and
      Goyal, Naman  and
      Chaudhary, Vishrav  and
      Wenzek, Guillaume  and
      Guzm{\'a}n, Francisco  and
      Grave, Edouard  and
      Ott, Myle  and
      Zettlemoyer, Luke  and
      Stoyanov, Veselin",
    booktitle = "Proceedings of the 58th Annual Meeting of the Association for Computational Linguistics",
    month = jul,
    year = "2020",
    address = "Online",
    publisher = "Association for Computational Linguistics",
    url = "https://aclanthology.org/2020.acl-main.747",
    doi = "10.18653/v1/2020.acl-main.747",
    pages = "8440--8451",
}

@inproceedings{bert,
    title = "{BERT}: Pre-training of Deep Bidirectional Transformers for Language Understanding",
    author = "Devlin, Jacob  and
      Chang, Ming-Wei  and
      Lee, Kenton  and
      Toutanova, Kristina",
    booktitle = "Proceedings of the 2019 Conference of the North {A}merican Chapter of the Association for Computational Linguistics: Human Language Technologies, Volume 1 (Long and Short Papers)",
    month = jun,
    year = "2019",
    address = "Minneapolis, Minnesota",
    publisher = "Association for Computational Linguistics",
    url = "https://aclanthology.org/N19-1423",
    doi = "10.18653/v1/N19-1423",
    pages = "4171--4186",
}

@inproceedings{opennmt,
    title = "{O}pen{NMT}: Open-Source Toolkit for Neural Machine Translation",
    author = "Klein, Guillaume  and
      Kim, Yoon  and
      Deng, Yuntian  and
      Senellart, Jean  and
      Rush, Alexander",
    booktitle = "Proceedings of {ACL} 2017, System Demonstrations",
    month = jul,
    year = "2017",
    address = "Vancouver, Canada",
    publisher = "Association for Computational Linguistics",
    url = "https://aclanthology.org/P17-4012",
    pages = "67--72",
}

@inproceedings{oostdijk-etal-2002-experiences,
    title = "Experiences from the Spoken {D}utch Corpus Project",
    author = "Oostdijk, Nelleke  and
      Goedertier, Wim  and
      van Eynde, Frank  and
      Boves, Louis  and
      Martens, Jean-Pierre  and
      Moortgat, Michael  and
      Baayen, Harald",
    booktitle = "Proceedings of the Third International Conference on Language Resources and Evaluation ({LREC}{'}02)",
    month = may,
    year = "2002",
    address = "Las Palmas, Canary Islands - Spain",
    publisher = "European Language Resources Association (ELRA)",
    url = "http://www.lrec-conf.org/proceedings/lrec2002/pdf/98.pdf",
}

@InProceedings{pmlr-v80-shazeer18a,
  title = 	 {Adafactor: Adaptive Learning Rates with Sublinear Memory Cost},
  author =       {Shazeer, Noam and Stern, Mitchell},
  booktitle = 	 {Proceedings of the 35th International Conference on Machine Learning},
  pages = 	 {4596--4604},
  year = 	 {2018},
  editor = 	 {Dy, Jennifer and Krause, Andreas},
  volume = 	 {80},
  series = 	 {Proceedings of Machine Learning Research},
  month = 	 {10--15 Jul},
  publisher =    {PMLR}
}

@misc{vaswani,
  doi = {10.48550/ARXIV.1706.03762},
  
  url = {https://arxiv.org/abs/1706.03762},
  
  author = {Vaswani, Ashish and Shazeer, Noam and Parmar, Niki and Uszkoreit, Jakob and Jones, Llion and Gomez, Aidan N. and Kaiser, Lukasz and Polosukhin, Illia},
  
  title = {Attention Is All You Need},
  
  publisher = {arXiv},
  
  year = {2017},
  
  copyright = {arXiv.org perpetual, non-exclusive license}
}

@inproceedings{moses,
    title = "{M}oses: Open Source Toolkit for Statistical Machine Translation",
    author = "Koehn, Philipp  and
      Hoang, Hieu  and
      Birch, Alexandra  and
      Callison-Burch, Chris  and
      Federico, Marcello  and
      Bertoldi, Nicola  and
      Cowan, Brooke  and
      Shen, Wade  and
      Moran, Christine  and
      Zens, Richard  and
      Dyer, Chris  and
      Bojar, Ond{\v{r}}ej  and
      Constantin, Alexandra  and
      Herbst, Evan",
    booktitle = "Proceedings of the 45th Annual Meeting of the Association for Computational Linguistics Companion Volume Proceedings of the Demo and Poster Sessions",
    month = jun,
    year = "2007",
    address = "Prague, Czech Republic",
    publisher = "Association for Computational Linguistics",
    url = "https://aclanthology.org/P07-2045",
    pages = "177--180",
}

@ARTICLE{Aronoff:2007,
AUTHOR = {Aronoff, M. },
TITLE   = {{L}anguage (linguistics)},
YEAR    = {2007},
JOURNAL = {Scholarpedia},
VOLUME  = {2},
NUMBER  = {5},
PAGES   = {3175},
DOI     = {10.4249/scholarpedia.3175},
NOTE    = {revision \#121088}
}

@inproceedings{SEPP_SharedTask, 
    author = "Don Tuggener and Ahmad Aghaebrahimian",
    title = {{The Sentence End and Punctuation Prediction in NLG Text (SEPP-NLG) Shared Task 2021}},
    year = 2021,
    booktitle = {Proceedings of the Swiss Text Analytics Conference 2021},
}

@inproceedings{attia-etal-2014-gwu,
    title = "{GWU}-{HASP}: Hybrid {A}rabic Spelling and Punctuation Corrector",
    author = "Attia, Mohammed  and
      Al-Badrashiny, Mohamed  and
      Diab, Mona",
    booktitle = "Proceedings of the {EMNLP} 2014 Workshop on {A}rabic Natural Language Processing ({ANLP})",
    month = oct,
    year = "2014",
    address = "Doha, Qatar",
    publisher = "Association for Computational Linguistics",
    url = "https://aclanthology.org/W14-3620",
    doi = "10.3115/v1/W14-3620",
    pages = "148--154",
}

@inproceedings{glove,
    title = "{G}lo{V}e: Global Vectors for Word Representation",
    author = "Pennington, Jeffrey  and
      Socher, Richard  and
      Manning, Christopher",
    booktitle = "Proceedings of the 2014 Conference on Empirical Methods in Natural Language Processing ({EMNLP})",
    month = oct,
    year = "2014",
    address = "Doha, Qatar",
    publisher = "Association for Computational Linguistics",
    url = "https://aclanthology.org/D14-1162",
    doi = "10.3115/v1/D14-1162",
    pages = "1532--1543",
}

@inproceedings{sunkara-etal-2020-robust,
    title = "Robust Prediction of Punctuation and Truecasing for Medical {ASR}",
    author = "Sunkara, Monica  and
      Ronanki, Srikanth  and
      Dixit, Kalpit  and
      Bodapati, Sravan  and
      Kirchhoff, Katrin",
    booktitle = "Proceedings of the First Workshop on Natural Language Processing for Medical Conversations",
    month = jul,
    year = "2020",
    address = "Online",
    publisher = "Association for Computational Linguistics",
    url = "https://aclanthology.org/2020.nlpmc-1.8",
    doi = "10.18653/v1/2020.nlpmc-1.8",
    pages = "53--62",
}

@article{biobert,
    author = {Lee, Jinhyuk and Yoon, Wonjin and Kim, Sungdong and Kim, Donghyeon and Kim, Sunkyu and So, Chan Ho and Kang, Jaewoo},
    title = "{BioBERT: a pre-trained biomedical language representation model for biomedical text mining}",
    journal = {Bioinformatics},
    volume = {36},
    number = {4},
    pages = {1234-1240},
    year = {2019},
    month = {09},
    issn = {1367-4803},
    doi = {10.1093/bioinformatics/btz682},
    url = {https://doi.org/10.1093/bioinformatics/btz682},
    eprint = {https://academic.oup.com/bioinformatics/article-pdf/36/4/1234/32527770/btz682.pdf},
}

@inproceedings{lison-tiedemann-2016-opensubtitles2016,
    title = "{O}pen{S}ubtitles2016: Extracting Large Parallel Corpora from Movie and {TV} Subtitles",
    author = {Lison, Pierre  and
      Tiedemann, J{\"o}rg},
    booktitle = "Proceedings of the Tenth International Conference on Language Resources and Evaluation ({LREC}'16)",
    month = may,
    year = "2016",
    address = "Portoro{\v{z}}, Slovenia",
    publisher = "European Language Resources Association (ELRA)",
    url = "https://aclanthology.org/L16-1147",
    pages = "923--929",
}

@article{Vandeghinste_Bulte_2019, title={Linguistic proxies of readability: Comparing easy-to-read and regular newspaper Dutch}, volume={9}, url={https://www.clinjournal.org/clinj/article/view/97}, abstractNote={&amp;lt;p&amp;gt;The aim of this study is to identify linguistic proxies of readability in Dutch, i.e. those linguistic features that define text as being easy-to-read. To this end, we compare the Wablieft corpus (Vandeghinste et al. 2019) (Flemish easy-to-read newspaper archives) to articles that appeared in the regular Flemish newspaper De Standaard, using a wide range of lexical, syntactic and readability metrics. We test which of these metrics has the highest effect size and which combinations of metrics work best in a classification task predicting whether articles belong to Wablieft or De Standaard. The results indicate that the best linguistic proxy for readability is (not surprisingly) the average number of words per sentence. Traditional reading metrics score well, although the combination of the parameters constituting these metrics score better in logistic regression than the original metrics.&amp;lt;/p&amp;gt;}, journal={Computational Linguistics in the Netherlands Journal}, author={Vandeghinste, Vincent and Bulté, Bram}, year={2019}, month={Dec.}, pages={81–100} }

\section*{Appendix}
\label{appendix}

In this appendix we present more detailed classification evaluation results, including precision and recall.


\begin{table}[H]
    \centering
    \begin{tabular}{r|rrrr}
    class &           precision& recall &   $F_1$-score&   samples\\
\hline
           0&   0.992584&  0.994595&  0.993588&   9627605\\
           .&   0.960450&  0.962452&  0.961450&    433554\\
           ,&   0.816974&  0.804882&  0.810883&    379759\\
           ?&   0.871368&  0.826812&  0.848506&     13494\\
           -&   0.619905&  0.367690&  0.461591&     27341\\
           :&   0.718636&  0.602076&  0.655212&     18305\\
\hline
    accuracy&&&                       0.983874 & 10500058\\
   macro avg &  0.829986&  0.759751&  0.788538&  10500058\\
weighted avg &  0.983302&  0.983874&  0.983492&  10500058\\
    \end{tabular}
    \caption{ Monolingual Europarl model tested on Nl EuroParl data.}
    \label{tab:ep_ep}
\end{table}


\begin{table}[H]
    \centering
    \begin{tabular}{r|rrrr}
        class   &precision &   recall&   $F_1$-score&   samples\\
\hline
0 &  0.982554 & 0.989277 &  0.985904 &  73926815 \\
. &  0.858432 & 0.852403 &  0.855407 &   4941897 \\
, &  0.754981 & 0.689276 &  0.720634 &   3127454 \\
? &  0.732037 & 0.646400 &  0.686558 &    410416 \\
- &  0.849020 & 0.629105 &  0.722703 &    331849 \\
: &  0.740604 & 0.659131 &  0.697497 &    590946 \\
\hline
accuracy  &          &        &   0.964436 & 83329377 \\
macro avg  & 0.819604 & 0.744266 & 0.778117 &  83329377 \\
weighted avg  & 0.963170 & 0.964436 & 0.963641 & 83329377 \\
    \end{tabular}
    \caption{Monolingual SoNaR model tested on Nl SoNaR.}
    \label{tab:ep_sonar}
\end{table}



\begin{table}[H]
    \centering
    \begin{tabular}{r|rrrr}
         class    &precision&    recall&   $F_1$-score&   samples\\
        \hline
        0 &  0.992625 & 0.994700 & 0.993662 &  9627605 \\
. &  0.960790 & 0.956852 & 0.958817 &   433554 \\
, &  0.815222 & 0.810991 & 0.813101 &   379759 \\
? &  0.867011 & 0.772047 & 0.816778 &    13494 \\
- &  0.657312 & 0.358070 & 0.463597 &    27341 \\
: &  0.708049 & 0.613166 & 0.657201 &    18305 \\
\hline
accuracy  &          &         &  0.983884 & 10500058 \\
macro avg &  0.833501 & 0.750971 & 0.783859 & 10500058 \\
weighted avg &  0.983364 & 0.983884 & 0.983499 & 10500058 \\
\end{tabular}
\caption{Multilingual EP model tested on Nl Europarl data}
\label{tab:multi}
\end{table}





\begin{table}[H]
    \centering
    \begin{tabular}{r|cccr}
   class  &precision&    recall&   $F_1$-score&   samples\\
         \hline
        0&   0.983286&  0.990781&  0.987020&   8982463\\
        .&   0.900062&  0.812584&  0.854089&    588253\\
       ,&   0.713272&  0.732957&  0.722980&    356718\\
       ?&   0.739526&  0.614814&  0.671428&     59631\\
       -&   0.727932&  0.529030&  0.612744&     32828\\              
       :&   0.725112&  0.694275&  0.709358&     49708\\       
\hline
    accuracy&&&                       0.966042&  10069601\\
   macro avg&   0.798198&  0.729074&  0.759603&  10069601\\
weighted avg&   0.965309&  0.966042&  0.965441&  10069601\\
\end{tabular}
\caption{Multilingual EP+SoNaR tested on Nl, Europarl + Nl SoNaR data.}
\end{table}

\end{document}